\documentclass[runningheads]{llncs}
\usepackage{graphicx}
\usepackage{multirow}
\usepackage{hyperref}
\graphicspath{ {./images/} }
\usepackage{adjustbox}
\begin{document}
{\let\thefootnote\relax\footnotetext{Copyright \textcopyright\ 2021 for this paper by its authors. Use permitted under Creative Commons License Attribution 4.0 International (CC BY 4.0).}}

\title{Multi-task Learning for Cross-Lingual Sentiment Analysis
}
%

%

\author{Gaurish Thakkar\orcidID{0000-0002-8119-5078}  \and
Nives Mikelic Preradović\orcidID{0000-0001-9087-0074} \and
Marko Tadić\orcidID{0000-0001-6325-820X}
}
\authorrunning{G. Thakkar et al.}
%

\institute{Faculty of Humanities and Social Sciences, University of Zagreb, Zagreb 10000, Croatia \\ \email{gthakkar@m.ffzg.hr}, nmikelic@m.ffzg.hr, marko.tadic@ffzg.hr}

\maketitle              
\begin{abstract}

This paper presents a cross-lingual sentiment analysis of news articles using zero-shot and few-shot learning. The study aims to classify the Croatian news articles with positive, negative, and neutral sentiments using the Slovene dataset. The system is based on a trilingual BERT-based model trained in three languages: English, Slovene, Croatian. The paper analyses different setups using datasets in two languages and proposes a simple multi-task model to perform sentiment classification. The evaluation is performed using the few-shot and zero-shot scenarios in single-task and multi-task experiments for Croatian and Slovene. 

\end{abstract}
%
%
%
\section{Introduction}
Sentiment analysis is one of the most exciting applications of Natural Language Processing. This field encompasses text analysis ranging from the average customer reviews of online products and movies to user-generated text from social media platforms. Other applications include the analysis of financial news \cite{agic2010towards,day2016deep} for stock market movement. While the field has a vast span across multiple subareas, here we are interested in sentiment analysis of news articles. This paper focuses on improving sentiment analysis of Croatian news articles tagged with coarse-grained sentiment tags. Since Slovene and Croatian belong to the same (sub)family of Slavic languages, Slovene was chosen as the hub language and in the context of rather a small dataset available for Croatian.

We seen the following items as the main contributions of our work:\textit{a) the mutual dependence of the sentiment analysis task at three levels (i.e., document-level, paragraph-level, and sentence-level) is leveraged to improve the performance of cross-lingual sentiment analysis; b) shared language encoder representations across both languages are proposed and c) a language from the same language family is used for cross-lingual sentiment transfer\footnote{Source code: https://github.com/cleopatra-itn/SentimentAnalyserSLHRNews}.}



\section{Related Work}
The previous state-of-the-art computational processes for sentiment analysis relied on sentiment lexicons \cite{taboada2011lexicon} as well as other classical methods like TF-IDF \cite{lin2008emotion} and dealing with various features along with SVM \cite{pang-etal-2002-thumbs}. As automatic feature extraction became a common trend with the usage of deep learning techniques \cite{socher2013recursive,tai2015improved}, the Convolutional Neural Nets \cite{wang-etal-2018-personalized} took the lead, and then gradually have been replaced by various Recurrent Neural Networks approaches \cite{dong2014adaptive,sutskever2014sequence,zadeh2018multi,majumder2019dialoguernn}.

The machine-translation, as one of the well-studied cross-lingual techniques, also aids to sentiment labelling. Its application ranges from lexicon translation \cite{abdalla-hirst-2017-cross,balahur2010sentiment} up to instance translations \cite{wan2009co}.
Several recent works have explored the use of transformers in multi-task learning setup for mono-lingual \cite{cer-etal-2018-universal} and multi-lingual setups \cite{chidambaram2018learning,DBLP:journals/corr/abs-1907-04307} for sentiment classification. 

The closest work upon which our research is built \cite{pelicon2020zero} performs zero-shot sentiment learning on Croatian news articles by enriching a masked language model (mBERT) \cite{devlin2018bert} using sentiment tags from the SentiNews and Croatian news sentiment dataset. Our work is novel because of the way we use the dataset. Previous work does not utilise paragraph-level and sentence-level annotations for document-level classification but uses them for pre-training a masked-language model. We utilise these annotations in a multi-task setup for aiding the sentiment classification task for Croatian. 


\section{Datasets}
We use two datasets in our experiments.

\textbf{SentiNews Dataset in Slovene}
This is a manually annotated dataset \cite{buvcar2018annotated}\footnote{https://www.clarin.si/repository/xmlui/handle/11356/1110} in the domain of news. It contains 10,427 documents. All annotations have three levels of granularity, i.e., document, paragraph, and sentence-level. The dataset covers news from economics, finance, and politics published between 1 September 2007 and 31 December 2013. It contains all instance annotations of each annotator, along with the news content and the final sentiment label. The dataset has been annotated at a five-point Likert scale and has mapping onto three labels (Positive, Negative, and Neutral) using the scale’s average score. The overall distribution of this dataset is given in Table 1. 
\begin{table}
\centering
\caption{Slovene and Croatian dataset statistics.}\label{tab2}
\begin{tabular}{|cccccc|}
\hline
\textbf{Lang}                     & \textbf{Level}      & \textbf{Examples} & \textbf{Positive} & \textbf{Negative} & \textbf{Neutral} \\ \hline
\multirow{3}{*}{Slovene} & Documents  & 10,427   & 1,665     & 3,337     & 5,425    \\
                         & Paragraphs & 89,999   & 14,636    & 23,721    & 51,642   \\ 
                         & Sentences  & 165,071   & 27,091    & 44,629    & 93,351   \\ \hline
Croatian                 & Documents  & 2,025     & 325      & 456      & 1,244    \\ \hline
\end{tabular}
\end{table}

\textbf{Sentiment Dataset in Croatian}
The Croatian dataset \footnote{https://www.clarin.si/repository/xmlui/handle/11356/1342} was created using guidelines similar to those of the SentiNews dataset. The text content comes from the Croatian 24sata daily news portal. It covers topics such as health, lifestyle, and automotive news. Table 1 shows the statistics for this dataset. Like the Slovenian corpus, this dataset is also annotated with 3-class sentiment labels and covers the same domain. However, it does not contain paragraph and sentence-level annotations. Both datasets present an imbalanced-class distribution phenomenon.

\section{Methodology}
We strive to leverage contextual information from the Slovene dataset accessible at three distinct levels and dataset from Croatian in our proposed system to promote knowledge transfer between two languages. As Croatian and Slovene proved to have the highest level of mutual intelligibility among three South Slavic languages (Croatian, Slovene and Bulgarian) represented in the mutual intelligibility study \cite{golubovic2015mutual}, 
we hypothesise Slovene could be the right candidate as a hub language for cross-lingual knowledge transfer to Croatian. Also, both datasets belong to the same text type – news. A simple multi-task learning setup with three different task heads is employed. Here, each task head represents the classification layer responsible for classifying the given instance of a particular type. The instances are of three types, namely document-level, paragraph-level, and sentence-level. The document-classification layer is trained by concatenating  the Croatian and Slovene dataset. The paragraph-classification and the sentence-classification layer are only fed with Slovene instances since the Croatian dataset does not provide this information. 

A shared encoder is used for feature extraction, enabling feature sharing across all three tasks and both languages. Our model is trained in five different scenarios in total, while the overall architecture is presented in Fig. 1. In the zero-shot learning setting, we do not use the Croatian data in training. However, use the Croatian test for reporting the performance.

\begin{enumerate}
    \item \textbf{Single-task-SL-Zero-shot-HR} - Train only Slovene document-level data. In this setting, the Slovene data is used in the zero-shot setting for the Croatian sentiment analysis. 
    \item \textbf{Multi-task-SL-Zero-shot-HR} - Train using all three levels on Slovene data. This setting is the same as the previous one, except that all three classification heads were learning respective tasks using Slovene data. We did not use any Croatian data in this or previous setting. 
    \item \textbf{Single-task-HR} - Train only using Croatian document-level data. This setting is a classic fine-tuning setup involving a single head learning classification on a single dataset. 
    \item \textbf{Multi-task-HR+SL} - Train with Croatian (document-level) and Slovene data (all three levels). We concatenated the respective document-level instances from respective languages at the same time. The other heads were trained using compatible instances from Slovene datasets. 
    \item \textbf{Single-task-HR+SL} - Train with Slovene and Croatian data (document-level). This approach is similar to the previous one, but the single document-level classification head is trained. 
\end{enumerate}

\begin{figure}
\centering

\includegraphics[scale=0.5]{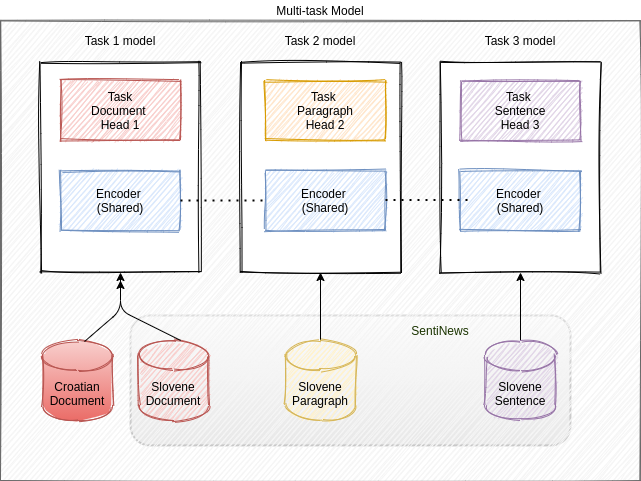}
\caption{The overall architecture of the proposed system. A shared encoder and three specific classification heads.}\label{fig1}
\end{figure}

\section{Experimental Setup}
This section presents a brief description of the pre-processing steps, followed by the details of the experiments.

\subsection{Preprocessing}
A few pre-processing steps were performed on the dataset. 
\begin{enumerate}
    \item The empty string values, which have neutral tags, were dropped. As null strings have no content, we decided to drop these instances.
    \item The strings based on content were de-duplicated. Many strings in the dataset were duplicates. We performed this step in order to prevent leakage of the instances into validation or test set split.
\end{enumerate}
Table 2 depicts the overall distribution of the dataset. There is a drop of 3k instances in the sentence-level Slovene dataset. 
 
\begin{table}
\caption{Dataset statistics after preprocessing.}\label{tab3}
\centering
\begin{tabular}{|l|l|r|r|r|r|}
\hline
\textbf{Language}                     & \textbf{Level}      & \textbf{Examples} & \textbf{Positive} & \textbf{Negative} & \textbf{Neutral} \\ \hline
\multirow{3}{*}{Slovene} & Documents  & 10,417    & 1,665     & 3,337     & 5,418    \\
                         & Paragraphs & 86,803    & 14,270    & 23,265    & 49,268   \\
                         & Sentences  & 161,291   & 26,679    & 44,014    & 90,598   \\ \hline
Croatian                 & Documents  & 1,988     & 321      & 450      & 1,217    \\ \hline
\end{tabular}

\end{table}

\subsection{Language Model - Shared encoder}
Our pipeline used a shared encoder based on a BERT-based masked language model. The tri-lingual model named \textit{CroSloEngual} \cite{10.1007/978-3-030-58323-1_11} trained with three languages (Slovene, Croatian, and English) on 5.9 billion tokens altogether. This model was chosen as it considers two languages that we are interested in transferring knowledge. The model outperforms the mBERT \cite{devlin2018bert} for the task of Part of Speech (POS) tagging, Named Entity Recognition (NER) and dependency parsing for these three languages. 

\subsection{Experiments}
In the datasets defined in Table 2, the data is split into 80:20 train-test split in a stratified fashion.
The 10\% or a proportionate train set is kept aside as a development set, since all the datasets differ in size. All datasets are combined in a single collection. The combined datasets behave as our size-proportional population. Our model is trained sequentially in a single-task setting by randomly sampling tasks from this collection. Using the test set from Croatian and Slovene, we evaluated our proposed approach. Table 3 shows the model configuration. All our models were trained using Nvidia RTX 3090 (24GB) with a batch size of 32. All hyperparameters were constant during the whole experiments except epoch which varied from 5 for a single task to 3 for a multi-task setup. These values were chosen to prevent overfitting on the train set. The overall training time for the MTL setup was 3 hours. We evaluated the performance on the development set and chose the best model for reporting test performance.

Table 3 reports all the results of our experiment. 
\begin{table}
\caption{Model Configurations.}\label{tab4}
\centering
\begin{tabular}{|c|c|}
\hline
\textbf{Parameters} & \textbf{Values}\\
\hline
Optimizer &   Adam (lr =2e-5)\\
Loss &  Categorical cross-entropy \\
Output  & Softmax \\
Batch & 32 \\
Epochs & 3-STL \& 5-MTL \\
Dropout & 0.3  \\
Labels & 3 \\
\hline
\end{tabular}
\end{table}

\subsection{Results}
The single-task (STL) results and the multi-task (MTL) setup results are reported in Table 4. Precision, recall, and F1 are macro averaged for all the experiments. We used a simple majority-class classifier as our baseline. Our MTL setup with Croatian and Slovene dataset outperforms the other settings for Croatian sentiment classification. However, it does not perform best when tested on document-level classification for Slovene. There is a small drop in the performance similar to what was previously reported by \cite{pelicon2020zero}. The second-best performing model for Croatian is the single-task variant with Slovene and Croatian data. The worst performing model on Croatian was the MTL with only Slovene data and the STL model, which was performing zero-shot learning. Nevertheless, the SL MTL performs better on Slovene test-sets. The HR STL, which used the least amount of data compared to other settings, seemed to perform at par with SL STL in F1(55.61 vs 56.95), but both had contrasting precision and recall. 

For paragraph-level, SL MTL has similar performance to SL+HR MTL, but we see a slight drop in performance in the latter case. A similar observation can be made for sentence-level classification. 


\begin{table}

\caption{Results of the experiments. Since the datasets are imbalanced, we present macro-averaged precision, recall, and F1 measure.}\label{tab5}
\begin{adjustbox}{width=\columnwidth,center}
\begin{tabular}{|l|c|c|c|c|c|c|c|c|c|c|c|c|}
\hline
\multicolumn{1}{|c|}{\textbf{Train set}} & \multicolumn{12}{c|}{\textbf{Test set}}                                                                                                           \\ \hline
\multirow{3}{*}{}                        & \multicolumn{9}{c|}{\textbf{Slovene}}                                                                   & \multicolumn{3}{c|}{\textbf{Croatian}}                    \\ \cline{2-13} 
                                         & \multicolumn{3}{c|}{\textbf{Document}} & \multicolumn{3}{c|}{\textbf{Paragraph}} & \multicolumn{3}{c|}{\textbf{Sentence}} & \multicolumn{3}{c|}{\textbf{Document}}                    \\ \cline{2-13} 
                                         & \textbf{P}        & \textbf{R}        & \textbf{F1}       & \textbf{P}        & \textbf{R}        & \textbf{F1}        & \textbf{P}        & \textbf{R}        & \textbf{F1}       & \textbf{P}              & \textbf{R}              & \textbf{F1}              \\ \hline
Majority class                                    & 17.33    & 33.33    & 22.80   & 18.91        & 33.33        & 24.13        & 18.72        & 33.33        & 23.97       & 20.43          & 33.33          & 25.33          \\ \hline
SL STL                                   & 70.56    & 71.65    & 71.07   & -        & -        & -        & -        & -        & -       & 57.88          & 63.86          & 55.61          \\ \hline
SL MTL                                   & \textbf{73.15}    & 77.66    & \textbf{74.86}   & \textbf{70.58}    & \textbf{73.50}    & 71.83    & \textbf{68.77}   & 70.13    & \textbf{69.40}   & 56.48    & 62.83          & 50.07          \\ \hline
HR STL                                   & 56.84    & 44.92    & 43.86   & -        & -        & -        & -        & -        & -       & 61.41          & 56.34          & 56.95          \\ \hline
SL+HR STL                                & 68.99    & 72.14    & 70.16   & -        & -        & -        & -        & -        & -       & 62.80          & 64.66          & 63.53          \\ \hline
SL+HR MTL                               & 72.32    & \textbf{78.00}   & 74.21   & 70.36    & 74.03    & \textbf{71.84}    & 68.32    & \textbf{70.34}    & 69.21   & \textbf{63.01} & \textbf{67.54} & \textbf{63.86} \\ \hline
\end{tabular}
\end{adjustbox}
\end{table}


\section{Discussion}
The research presented in this paper has revealed that having a small amount of target language (Croatian) data helps in overall cross-lingual transfer learning. Adding data from another language hinders the source language task’s performance but improves task performance in the target language. 
Comparing our work with \cite{pelicon2020zero}, which proposes joint optimisation of language modelling task and paragraph-level sentiment analysis for the document-level sentiment classification, we did not perform any intermediate training but utilised the available annotation directly at the fine-tuning stage. Another significant difference is the use of a trilingual shared encoder which performs better than mBERT.  
We have qualitatively analysed the MTL model’s errors on the test set and discovered that the model does pick up cues from the sentiment bearing words from the input. However, the analysis of these errors would be beyond the scope of this paper. The current model tags the news document as positive or negative when it finds positive or negative words in neutral news. The articles, which are advertisements or recipes, contain words with positive sentiment. Most of the errors belong to this class. Since our encoder has fixed-length input, the truncated text prevents correct classification. We can solve this problem by performing a sliding window sampling over the text or using the beginning and the end of each article.  

\section{Conclusion}
We presented an overall setup for cross-lingual sentiment analysis (SA) of Croatian news documents using a Multi-task learning approach. The goal was to perform knowledge transfer using existing datasets and models to aid SA in Croatian. For this purpose, we used a large Slovene SentiNews corpus created similarly to the Croatian corpus. 

The publicly available trained trilingual BERT-based language model for feature representation was utilised and used as a shared encoder for the downstream tasks. We combined the Croatian and Slovene datasets for document-level classification and trained three different classification heads. The results show that the MTL setup outperformed the STL setup for Croatian. 

In the future, we would like to experiment more in an under-resourced setting for Slovene and balance the dataset among distinct levels. Another interesting approach would be to use each level of the datasets individually in a hierarchical fashion to help with the next level's classification task. For example, using sentence-level features to aid the paragraph classification could be fused into the document-level prediction process. We believe that processing documents from different text types and topics separately would be an easier task. Therefore, in our future work, we plan to cluster documents into text-types and topics like recipes, advertisements or obituaries and process them separately. 
Also, our experiment setting could be further checked by running it for other language pairs supported by this \textit{CroSloEngual} BERT-based model: English-Slovene and English-Croatian. In this way, we could verify whether the genetically and typologically distant language, as English here is, would contribute to the performance.  

\section{Acknowledgements}
The work presented in this paper has received funding from the European Union’s Horizon 2020 research and innovation program under the Marie Skłodowska-Curie grant agreement no. 812997 and under the name CLEOPATRA (Cross-lingual Event-centric Open Analytics Research Academy).
%
%
%
\bibliographystyle{splncs04}
\bibliography{bibliography}

\begin{thebibliography}{10}
\providecommand{\url}[1]{\texttt{#1}}
\providecommand{\urlprefix}{URL }
\providecommand{\doi}[1]{https://doi.org/#1}

\bibitem{abdalla-hirst-2017-cross}
Abdalla, M., Hirst, G.: Cross-lingual sentiment analysis without (good)
  translation. In: Proceedings of the Eighth International Joint Conference on
  Natural Language Processing (Volume 1: Long Papers). pp. 506--515. Asian
  Federation of Natural Language Processing, Taipei, Taiwan (Nov 2017),
  \url{https://www.aclweb.org/anthology/I17-1051}

\bibitem{agic2010towards}
Agi{\'c}, {\v{Z}}., Ljube{\v{s}}i{\'c}, N., Tadi{\'c}, M.: Towards sentiment
  analysis of financial texts in croatian. In: Proceedings of the Seventh
  International Conference on Language Resources and Evaluation (LREC'10)
  (2010)

\bibitem{balahur2010sentiment}
Balahur, A., Steinberger, R., Kabadjov, M., Zavarella, V., van~der Goot, E.,
  Halkia, M., Pouliquen, B., Belyaeva, J.: Sentiment analysis in the news. In:
  Proceedings of the Seventh International Conference on Language Resources and
  Evaluation (LREC'10) (2010)

\bibitem{buvcar2018annotated}
Bu{\v{c}}ar, J., {\v{Z}}nidar{\v{s}}i{\v{c}}, M., Povh, J.: Annotated news
  corpora and a lexicon for sentiment analysis in slovene. Language Resources
  and Evaluation  \textbf{52}(3),  895--919 (2018)

\bibitem{cer-etal-2018-universal}
Cer, D., Yang, Y., Kong, S.y., Hua, N., Limtiaco, N., St.~John, R., Constant,
  N., Guajardo-Cespedes, M., Yuan, S., Tar, C., Strope, B., Kurzweil, R.:
  Universal sentence encoder for {E}nglish. In: Proceedings of the 2018
  Conference on Empirical Methods in Natural Language Processing: System
  Demonstrations. pp. 169--174. Association for Computational Linguistics,
  Brussels, Belgium (Nov 2018). \doi{10.18653/v1/D18-2029},
  \url{https://www.aclweb.org/anthology/D18-2029}

\bibitem{chidambaram2018learning}
Chidambaram, M., Yang, Y., Cer, D., Yuan, S., Sung, Y.H., Strope, B., Kurzweil,
  R.: Learning cross-lingual sentence representations via a multi-task
  dual-encoder model. arXiv preprint arXiv:1810.12836  (2018)

\bibitem{day2016deep}
Day, M.Y., Lee, C.C.: Deep learning for financial sentiment analysis on finance
  news providers. In: 2016 IEEE/ACM International Conference on Advances in
  Social Networks Analysis and Mining (ASONAM). pp. 1127--1134. IEEE (2016)

\bibitem{devlin2018bert}
Devlin, J., Chang, M.W., Lee, K., Toutanova, K.: Bert: Pre-training of deep
  bidirectional transformers for language understanding. arXiv preprint
  arXiv:1810.04805  (2018)

\bibitem{dong2014adaptive}
Dong, L., Wei, F., Tan, C., Tang, D., Zhou, M., Xu, K.: Adaptive recursive
  neural network for target-dependent twitter sentiment classification. In:
  Proceedings of the 52nd annual meeting of the association for computational
  linguistics (volume 2: Short papers). pp. 49--54 (2014)

\bibitem{golubovic2015mutual}
Golubovi{\'c}, J., Gooskens, C.: Mutual intelligibility between west and south
  slavic languages. Russian linguistics  \textbf{39}(3),  351--373 (2015)

\bibitem{lin2008emotion}
Lin, K.H.Y., Yang, C., Chen, H.H.: Emotion classification of online news
  articles from the reader's perspective. In: 2008 IEEE/WIC/ACM International
  Conference on Web Intelligence and Intelligent Agent Technology. vol.~1, pp.
  220--226. IEEE (2008)

\bibitem{majumder2019dialoguernn}
Majumder, N., Poria, S., Hazarika, D., Mihalcea, R., Gelbukh, A., Cambria, E.:
  Dialoguernn: An attentive rnn for emotion detection in conversations. In:
  Proceedings of the AAAI Conference on Artificial Intelligence. vol.~33, pp.
  6818--6825 (2019)

\bibitem{pang-etal-2002-thumbs}
Pang, B., Lee, L., Vaithyanathan, S.: Thumbs up? sentiment classification using
  machine learning techniques. In: Proceedings of the 2002 Conference on
  Empirical Methods in Natural Language Processing ({EMNLP} 2002). pp. 79--86.
  Association for Computational Linguistics (Jul 2002).
  \doi{10.3115/1118693.1118704},
  \url{https://www.aclweb.org/anthology/W02-1011}

\bibitem{pelicon2020zero}
Pelicon, A., Pranji{\'c}, M., Miljkovi{\'c}, D., {\v{S}}krlj, B., Pollak, S.:
  Zero-shot learning for cross-lingual news sentiment classification. Applied
  Sciences  \textbf{10}(17), ~5993 (2020)

\bibitem{socher2013recursive}
Socher, R., Perelygin, A., Wu, J., Chuang, J., Manning, C.D., Ng, A.Y., Potts,
  C.: Recursive deep models for semantic compositionality over a sentiment
  treebank. In: Proceedings of the 2013 conference on empirical methods in
  natural language processing. pp. 1631--1642 (2013)

\bibitem{sutskever2014sequence}
Sutskever, I., Vinyals, O., Le, Q.V.: Sequence to sequence learning with neural
  networks. arXiv preprint arXiv:1409.3215  (2014)

\bibitem{taboada2011lexicon}
Taboada, M., Brooke, J., Tofiloski, M., Voll, K., Stede, M.: Lexicon-based
  methods for sentiment analysis. Computational linguistics  \textbf{37}(2),
  267--307 (2011)

\bibitem{tai2015improved}
Tai, K.S., Socher, R., Manning, C.D.: Improved semantic representations from
  tree-structured long short-term memory networks. In: Proceedings of the 53rd
  Annual Meeting of the Association for Computational Linguistics and the 7th
  International Joint Conference on Natural Language Processing (Volume 1: Long
  Papers). pp. 1556--1566 (2015)

\bibitem{10.1007/978-3-030-58323-1_11}
Ul{\v{c}}ar, M., Robnik-{\v{S}}ikonja, M.: Finest bert and crosloengual bert.
  In: Sojka, P., Kope{\v{c}}ek, I., Pala, K., Hor{\'a}k, A. (eds.) Text,
  Speech, and Dialogue. pp. 104--111. Springer International Publishing, Cham
  (2020)

\bibitem{wan2009co}
Wan, X.: Co-training for cross-lingual sentiment classification. In:
  Proceedings of the Joint Conference of the 47th Annual Meeting of the ACL and
  the 4th International Joint Conference on Natural Language Processing of the
  AFNLP. pp. 235--243 (2009)

\bibitem{wang-etal-2018-personalized}
Wang, W., Feng, S., Gao, W., Wang, D., Zhang, Y.: Personalized microblog
  sentiment classification via adversarial cross-lingual multi-task learning.
  In: Proceedings of the 2018 Conference on Empirical Methods in Natural
  Language Processing. pp. 338--348. Association for Computational Linguistics,
  Brussels, Belgium (Oct-Nov 2018). \doi{10.18653/v1/D18-1031},
  \url{https://www.aclweb.org/anthology/D18-1031}

\bibitem{DBLP:journals/corr/abs-1907-04307}
Yang, Y., Cer, D., Ahmad, A., Guo, M., Law, J., Constant, N., {\'{A}}brego,
  G.H., Yuan, S., Tar, C., Sung, Y., Strope, B., Kurzweil, R.: Multilingual
  universal sentence encoder for semantic retrieval. CoRR
  \textbf{abs/1907.04307} (2019), \url{http://arxiv.org/abs/1907.04307}

\bibitem{zadeh2018multi}
Zadeh, A., Liang, P.P., Poria, S., Vij, P., Cambria, E., Morency, L.P.:
  Multi-attention recurrent network for human communication comprehension. In:
  Proceedings of the AAAI Conference on Artificial Intelligence. vol.~32 (2018)

\end{thebibliography}

\end{document}